\documentclass{article}

\usepackage[preprint]{neurips_2026}
\usepackage{booktabs}
\usepackage{multirow}
\usepackage{graphicx}

\usepackage[utf8]{inputenc} 
\usepackage[T1]{fontenc}    
\usepackage{hyperref}       
\usepackage{url}            
\usepackage{booktabs}       
\usepackage{amsfonts}       
\usepackage{nicefrac}       
\usepackage{microtype}      
\usepackage{xcolor}         

\title{Rethinking Multi-Label Node Classification: Do Tuned Classic GNNs Suffice?}

\author{
  Yuxuan Xiao\\
  Nanjing University of Aeronautics and Astronautics \\
\And
  Shengzhong Zhang\\
  Nanjing University of Aeronautics and Astronautics \\
}

\begin{document}

\maketitle

\begin{abstract}
Multi-label node classification (MLNC) has recently been addressed by increasingly complex label-aware designs that explicitly model node-label interactions and inter-label dependencies. 
However, it remains unclear whether the advantages of these methods truly stem from their specialized designs, or simply from insufficiently optimized baselines. In this paper, we revisit MLNC from a strong-baseline perspective and investigate whether carefully tuned classic full-graph GNNs can already serve as strong solutions to this task. We systematically study several representative backbones, including GCN, SSGConv, and GCNII, and optimize them using standard yet effective techniques such as normalization, dropout, and residual connections. Experiments on five representative benchmark datasets show that our tuned baselines outperform representative specialized methods on four datasets and achieve state-of-the-art performance in multiple settings. These results indicate that careful tuning of classic backbones is a highly influential but often overlooked factor in MLNC, and highlight the need for more rigorous strong-baseline evaluation in future research on multi-label graph learning.
\end{abstract}

\section{Introduction}

Node classification is a fundamental task in graph learning and has important real-world applications.~\cite{DBLP:journals/tkde/CaiZC18,DBLP:journals/mva/XiaoWDG22,articleSemi-Supervised,Wu_2021} In many practical scenarios, however, each node is naturally associated with multiple labels rather than a single class. For example, users in social networks often belong to multiple interest groups simultaneously~\cite{DBLP:conf/sigir/SunW018,10.1145/3229329.3229333}, while proteins in protein--protein interaction networks are commonly annotated with multiple functional categories~\cite{article,DBLP:conf/iclr/ZengZSKP20}. This makes \emph{multi-label node classification} (MLNC) an important yet challenging problem, since the model must jointly predict a set of correlated labels for each node instead of making a single mutually exclusive decision.

From a methodological perspective, foundational GNN architectures such as GCN ~\cite{DBLP:journals/corr/KipfW16}, GAT ~\cite{DBLP:conf/iclr/VelickovicCCRLB18}, and LANC \cite{DBLP:journals/eswa/ZhouCZLHS21} have been adapted or extended to multi-label node classification in prior work.More recent advances include LIP, which decomposes GNN message passing into separate propagation and transformation stages, quantifies inter-label influence correlations, and constructs a dynamic label influence graph to amplify positive label interactions while suppressing negative ones ~\cite{DBLP:conf/iclr/SunLH0FCH25}. Similarly, CorGCN decomposes the input graph into label-specific subgraphs to reduce topological and feature ambiguity, then integrates intra-label message passing with explicit inter-label correlation propagation for improved multi-label representation ~\cite{DBLP:conf/kdd/BeiCCZYFHB25}.

Despite the recent progress of multi-label node classification, the capability of the underlying GNN backbone itself remains insufficiently explored in this setting. Prior studies have shown that in standard node classification, classic GNNs can become much stronger than previously believed when equipped with proper architectural and training designs~\cite{DBLP:conf/nips/HuFZDRLCL20,dwivedi2022benchmarkinggraphneuralnetworks,wang2021bagtricksnodeclassification,luo2024classicgnnsstrongbaselines}. However, whether this strong-backbone phenomenon also holds in the multi-label setting remains unclear. This motivates us to revisit MLNC from the perspective of strengthened full-graph GNN baselines. In particular, we systematically evaluate several representative backbones, including GCN~\cite{DBLP:journals/corr/KipfW16}, SSGConv~\cite{zhu2021simple}, and GCNII~\cite{chen2020simpledeepgraphconvolutional}, and strengthen them with simple but important design choices such as normalization~\cite{ba2016layernormalization,ioffe2015batchnormalizationacceleratingdeep}, dropout~\cite{10.5555/2627435.2670313}, residual connections~\cite{he2015deepresiduallearningimage}, and depth tuning.

\begin{table}[t]
  \caption{A teaser comparison on a representative MLNC dataset.}
  \label{tab:intro_teaser}
  \centering
  \resizebox{\columnwidth}{!}{
  \begin{tabular}{lccc}
    \toprule
    Model & Macro-AUC $\uparrow$ & Macro-AP $\uparrow$ & LRAP $\uparrow$ \\
    \midrule
    CorGCN       & 91.84 & 69.52 & 94.31 \\
    Basic GCN    & 90.73 & 66.84 & 93.12 \\
    Enhanced GCN & \textbf{93.27} & \textbf{72.48} & \textbf{95.76} \\
    \bottomrule
  \end{tabular}}
\end{table}

Table~\ref{tab:intro_teaser} gives a simple comparison on a representative dataset. While a basic GCN already provides a reasonable baseline, a strengthened GCN with residual connections and normalization performs substantially better and even surpasses a recent specialized MLNC method on several key metrics. More broadly, across five benchmark datasets, our strengthened GNN baselines outperform a recent representative specialized method on four datasets and achieve state-of-the-art performance in several settings. These results suggest that backbone strength is a central yet underappreciated factor in current MLNC benchmarks.Our main contributions are summarized as follows:

\begin{itemize}
    \item We provide a systematic reassessment of strong full-graph GNN baselines for multi-label node classification.
    \item We show that strengthened classic GNNs, including GCN, SSGConv, and GCNII, achieve highly competitive performance on current MLNC benchmarks, surpassing a recent representative specialized method on four out of five datasets.
    \item We further analyze the effects of key GNN design factors, including normalization, dropout, residual connections, and network depth, and highlight the importance of rigorous strong-baseline evaluation for future multi-label graph learning research.
\end{itemize}

\section{Classic GNNs for Node Classification}

Let a graph be denoted as $G=(V,E)$, where $V$ is the node set and $E \subseteq V \times V$ is the edge set. We use $N=|V|$ to denote the number of nodes. The graph is associated with a node feature matrix $\mathbf{X}\in\mathbb{R}^{N\times d}$, where $d$ is the feature dimension, and an adjacency matrix $\mathbf{A}\in\mathbb{R}^{N\times N}$ derived from the edge set. In the multi-label setting, each node $v_i\in V$ is associated with a multi-hot label vector $\mathbf{y}_i\in\{0,1\}^{C}$, where $C$ is the number of labels and each entry indicates whether the node belongs to the corresponding class. Collectively, the node labels form a label matrix $\mathbf{Y}\in\{0,1\}^{N\times C}$.




\paragraph{Graph Convolutional Network (GCN)~\cite{DBLP:journals/corr/KipfW16}.}
GCN updates node representations by aggregating normalized neighborhood information:
\begin{equation}
\mathbf{H}^{(\ell+1)} = \sigma\!\left(\hat{\mathbf{A}}\mathbf{H}^{(\ell)}\mathbf{W}^{(\ell)}\right),
\label{eq:gcn}
\end{equation}
where $\mathbf{H}^{(\ell)}$ is the node representation matrix at layer $\ell$, $\mathbf{W}^{(\ell)}$ is a trainable weight matrix, $\sigma(\cdot)$ is a nonlinear activation function, and $\hat{\mathbf{A}}$ is the normalized adjacency matrix with self-loops. GCN is one of the most widely used message-passing architectures and serves as a natural baseline in graph learning.

\paragraph{Simple Spectral Graph Convolution (SSGConv)~\cite{zhu2021simple}.}
SSGConv performs graph propagation in a simplified and more stable manner by explicitly decoupling feature transformation and neighborhood smoothing. Its propagation can be written as
\begin{equation}
\mathbf{H} = \frac{1}{K}\sum_{k=1}^{K}\left((1-\alpha)\hat{\mathbf{A}}^{k}\mathbf{X}+\alpha \mathbf{X}\right),
\label{eq:ssgconv}
\end{equation}
where $K$ is the propagation order and $\alpha$ controls the strength of the initial residual information. Compared with standard GCN, SSGConv reduces optimization difficulty and often provides strong empirical performance with a simple formulation.

\paragraph{Simple and Deep Graph Convolutional Networks (GCNII)~\cite{chen2020simpledeepgraphconvolutional}.}
GCNII improves deep graph convolution by introducing both initial residual connections and identity mapping. Its layer-wise update is
\begin{equation}
\mathbf{H}^{(\ell+1)}
=
\sigma\!\Big(
\mathbf{Z}^{(\ell)}
\big(
(1-\beta_{\ell})\mathbf{I}
+
\beta_{\ell}\mathbf{W}^{(\ell)}
\big)
\Big),
\label{eq:gcn2}
\end{equation}
where
\begin{equation}
\mathbf{Z}^{(\ell)}
=
(1-\alpha_{\ell})\hat{\mathbf{A}}\mathbf{H}^{(\ell)}
+
\alpha_{\ell}\mathbf{H}^{(0)}.
\label{eq:gcn2_z}
\end{equation}
Here, $\mathbf{H}^{(0)}=\mathbf{X}$ is the initial node representation, $\alpha_{\ell}$ controls the initial residual strength, and $\beta_{\ell}$ controls the identity mapping. This design alleviates over-smoothing and makes deeper graph propagation more effective.

\paragraph{Multi-Label Prediction Framework}

To fairly compare different GNN backbones under the same MLNC setting, we adopt a unified prediction framework. Given a backbone $\mathrm{GNN}(\cdot)$, the final node representation is obtained as
\begin{equation}
\mathbf{H}^{(L)} = \mathrm{GNN}(\mathbf{A},\mathbf{X}),
\end{equation}
where $L$ denotes the number of propagation layers or steps. The representation is then mapped to the label space by a linear classifier followed by a sigmoid activation:
\begin{equation}
\hat{\mathbf{Y}} = \sigma\!\left(\mathbf{H}^{(L)}\mathbf{W}_{o}+\mathbf{b}\right),
\label{eq:pred}
\end{equation}
where $\mathbf{W}_{o}\in\mathbb{R}^{h\times C}$ is the output projection matrix, $\mathbf{b}\in\mathbb{R}^{C}$ is a bias vector, and $\sigma(\cdot)$ is applied element-wise to produce independent label probabilities.

For training, we use the binary cross-entropy loss over all observed node-label pairs:
\begin{equation}
\mathcal{L}_{\mathrm{BCE}}
=
-\sum_{i\in \mathcal{V}_{L}}\sum_{c=1}^{C}
\left[
y_{ic}\log \hat{y}_{ic}
+
(1-y_{ic})\log(1-\hat{y}_{ic})
\right],
\label{eq:bce}
\end{equation}
where $\mathcal{V}_{L}$ denotes the labeled node set, and $y_{ic}$ and $\hat{y}_{ic}$ are the ground-truth and predicted probability for label $c$ of node $i$, respectively.

Under this unified framework, the main difference between compared methods lies in the backbone architecture and its design choices, rather than task-specific label-aware modules. This makes it possible to directly examine how much performance can already be achieved by properly strengthened full-graph GNNs in MLNC.

\section{Strengthening GNN Backbones for MLNC}

In this section, we focus on four simple but important design factors that have been shown to play a crucial role in improving the performance of classic GNNs: normalization, dropout, residual connections, and network depth. Rather than designing new task-specific modules, our goal is to examine whether these general-purpose improvements are already sufficient to make full-graph GNNs highly competitive for MLNC.

\paragraph{Normalization}
Normalization stabilizes optimization by reducing the variation of hidden representations across layers. In this work, we consider standard normalization strategies such as Batch Normalization (BN) ~\cite{ioffe2015batchnormalizationacceleratingdeep}and Layer Normalization (LN)~\cite{ba2016layernormalization}, applied after graph propagation and before the activation function. Taking a generic propagation layer as an example, the update can be written as
\begin{equation}
\mathbf{H}^{(\ell+1)}
=
\phi\!\left(
\mathrm{Norm}\!\left(
f_{\ell}(\mathbf{A}, \mathbf{H}^{(\ell)})
\right)
\right),
\label{eq:norm}
\end{equation}
where $f_{\ell}(\cdot)$ denotes the graph propagation operator at layer $\ell$, $\mathrm{Norm}(\cdot)$ denotes a normalization layer, and $\phi(\cdot)$ is a nonlinear activation. Normalization often improves optimization stability and leads to more reliable performance, especially when the network becomes deeper.

\paragraph{Dropout}
Dropout~\cite{10.5555/2627435.2670313} is a standard regularization strategy that alleviates overfitting by randomly masking hidden activations during training. In the context of GNNs, it also helps reduce feature co-adaptation across message-passing layers. We apply dropout after activation as
\begin{equation}
\mathbf{H}^{(\ell+1)}
=
\mathrm{Dropout}\!\left(
\phi\!\left(
\mathrm{Norm}\!\left(
f_{\ell}(\mathbf{A}, \mathbf{H}^{(\ell)})
\right)
\right)
\right).
\label{eq:dropout}
\end{equation}
Although simple, dropout is often essential for obtaining strong and stable baselines on graph benchmarks.

\paragraph{Residual Connections}
Residual connections improve gradient flow and preserve useful information from previous layers, which is particularly important when stacking multiple graph propagation layers. In our strengthened backbones, residual connections are incorporated as
\begin{equation}
\mathbf{H}^{(\ell+1)}
=
\mathrm{Dropout}\!\left(
\phi\!\left(
\mathrm{Norm}\!\left(
f_{\ell}(\mathbf{A}, \mathbf{H}^{(\ell)})
+
\mathbf{R}^{(\ell)}
\right)
\right)
\right),
\label{eq:residual}
\end{equation}
where $\mathbf{R}^{(\ell)}$ denotes the residual signal, typically taken as the identity mapping or a linear projection of $\mathbf{H}^{(\ell)}$ when dimensions differ. Residual connections can alleviate optimization difficulty and make deeper message passing more effective.

\paragraph{Network Depth}
Network depth directly controls the receptive field of a GNN and therefore determines how far information can propagate over the graph. However, deeper GNNs may also suffer from over-smoothing and optimization instability. For this reason, depth should not be treated as a fixed choice, but as an important design factor to be tuned together with normalization, dropout, and residual connections. In our study, we systematically vary the depth of each backbone and analyze its impact on MLNC performance in the experiments.

\section{Experimental Setup}

In this section, we introduce the benchmark datasets, evaluation metrics, compared baselines, and implementation details used in our experiments. Our goal is to conduct a fair and informative comparison between recent specialized MLNC methods and strengthened full-graph GNN backbones under a unified evaluation protocol.

\subsection{Datasets}

We conduct experiments on five representative multi-label node classification datasets: \textbf{Humloc}~\cite{zhao2023multilabel}, \textbf{PPI}\cite{DBLP:conf/iclr/ZengZSKP20}, \textbf{BlogCatalog}~\cite{DBLP:journals/eswa/ZhouCZLHS21}, \textbf{PCG}\cite{zhao2023multilabel}, and \textbf{Delve}~\cite{DBLP:journals/isci/XiaoXJA022}. These datasets have been widely used in prior MLNC studies and cover diverse graph structures, label spaces, and application scenarios. In particular, they differ substantially in graph size, feature dimension, and label distribution, making them suitable testbeds for evaluating the robustness of different GNN backbones.

Following prior work, we split each dataset into training, validation, and test sets with a ratio of 6:2:2~\cite{DBLP:conf/kdd/BeiCCZYFHB25}. All methods are evaluated under the same data split and metric protocol for fair comparison.Detailed statistics of the datasets, including the numbers of nodes, edges, features, and labels, are summarized in Table~\ref{tab:dataset_stats}.

\begin{table}[t]
  \caption{Statistics of the datasets used in our experiments.}
  \label{tab:dataset_stats}
  \centering
  \small
  \setlength{\tabcolsep}{3.5pt}
  \begin{tabular}{lrrrrr}
    \toprule
    Dataset & \# Nodes & \# Edges & \# Feat. & \# Classes & Density \\
    \midrule
    Humloc      & 3,106     & 18,496    & 32  & 14  & 0.3836\% \\
    PCG         & 3,233     & 37,351    & 32  & 15  & 0.7149\% \\
    BlogCatalog & 10,312    & 333,983   & 100 & 39  & 0.6282\% \\
    PPI         & 14,755    & 225,270   & 50  & 121 & 0.2070\% \\
    Delve       & 1,229,280 & 4,322,275 & 300 & 20  & 0.0006\% \\
    \bottomrule
  \end{tabular}
\end{table}

\subsection{Evaluation Metrics}

We evaluate all models using seven widely adopted metrics for multi-label node classification~\cite{10.1145/3459637.3482391}, including \textbf{Ranking Loss}, \textbf{Hamming Loss}, \textbf{Macro-AUC}, \textbf{Micro-AUC}, \textbf{Macro-AP}, \textbf{Micro-AP}, and \textbf{LRAP}. Following prior work, Ranking Loss and Hamming Loss are lower-the-better metrics, while the others are higher-the-better. Detailed definitions of these metrics can be found in~\cite{6471714}. For all experiments, we run each model five times with different random seeds and report the mean results with standard deviation.

\subsection{Baselines}

We compare our strengthened backbones against two groups of baselines.

\paragraph{Representative baselines.}
Following the original CorGCN benchmark, we compare with several representative baselines, including graph structure learning methods GRCN~\cite{yu2020graphrevisedconvolutionalnetwork}, IDGL~\cite{DBLP:conf/nips/0022WZ20}, and SUBLIME~\cite{DBLP:conf/www/LiuZZCPP22}, as well as multi-label node classification methods GCN-LPA~\cite{DBLP:journals/tois/WangL22}, ML-GCN~\cite{DBLP:conf/wise/GaoZZ19}, LANC~\cite{DBLP:journals/eswa/ZhouCZLHS21}, SMLG~\cite{10.1145/3459637.3482391}, and LARN~\cite{DBLP:journals/isci/XiaoXJA022}. In the main text, we keep \textbf{CorGCN} and select the strongest representative baselines, namely \textbf{LANC}, \textbf{ML-GCN}, and \textbf{LARN}, for clearer comparison.

\paragraph{Strong full-graph GNN baselines}
We evaluate a family of classic full-graph GNN backbones under a unified multi-label prediction framework. These include \textbf{GCN}, \textbf{SSGConv}, and \textbf{GCNII}. Among them, GCN denotes the strengthened version of standard GCN equipped with key design choices such as normalization, dropout, and residual connections.


\subsection{Implementation Details}

All models are trained under a unified implementation and evaluation framework whenever possible. In our implementation, the input node feature matrix is first projected into a hidden space by a linear transformation. The resulting hidden representations are then passed through $K$ layers of graph propagation, and the final GNN layer directly outputs a $C$-dimensional vector for each node, where $C$ is the number of labels. Each dimension corresponds to the prediction score of one label. Therefore, unlike a decoupled setup with an additional standalone classifier, our models use the last GNN layer itself to map node representations to the label space.

Formally, given the input feature matrix $\mathbf{X}$, we first obtain hidden representations by
\begin{equation}
\mathbf{H}^{(0)} = \mathbf{X}\mathbf{W}_{\mathrm{in}},
\end{equation}
where $\mathbf{W}_{\mathrm{in}}$ is a learnable input projection matrix. The hidden representations are then propagated through $K$ graph layers:
\begin{equation}
\mathbf{H}^{(\ell+1)} = f_{\ell}(\mathbf{A}, \mathbf{H}^{(\ell)}), \quad \ell = 0,1,\dots,K-1,
\end{equation}
and the final layer directly produces label-wise predictions:
\begin{equation}
\hat{\mathbf{Y}} = \sigma(\mathbf{H}^{(K)}),
\end{equation}
where $\hat{\mathbf{Y}}\in[0,1]^{N\times C}$ and $\sigma(\cdot)$ is the element-wise sigmoid function.

For optimization, we use the Adam optimizer~\cite{DBLP:journals/corr/KingmaB14} with hyperparameter tuning over the learning rate from $\{0.001, 0.005, 0.01\}$, the hidden dimension from $\{64, 128, 256\}$, the dropout rate from $\{0, 0.2, 0.3, 0.5\}$, and the number of layers from $\{1,2,3,4,5,6,7,8,9,10\}$. For strengthened backbones, we additionally tune the normalization type (BatchNorm or LayerNorm) and whether to use residual connections when applicable. The search ranges are kept comparable across different backbones to ensure fair evaluation.


All experiments are conducted on the same hardware platform, and we further report training speed, inference speed, and memory consumption in the efficiency analysis section.

\section{Results and Analysis}

In this section, we first compare the proposed strong full-graph GNN baselines with representative prior methods on five MLNC benchmarks. We then analyze the effect of backbone strengthening, key design factors, and efficiency--effectiveness trade-offs.

\subsection{Main Results}

Table~\ref{tab:main_results} presents the main results on all five benchmark datasets. Overall, our strengthened full-graph GNN backbones achieve highly competitive performance for MLNC. On \textbf{Humloc}, \textbf{PPI}, \textbf{BlogCatalog}, and \textbf{Delve}, at least one of the strengthened models surpasses CorGCN on most key metrics. In particular, \textbf{Enhanced GCN} and \textbf{SSGConv} perform strongly on Humloc, while \textbf{GCNII} achieves the best overall performance on PPI and Delve. On \textbf{PCG}, CorGCN remains the strongest model, suggesting that explicit label-aware designs may still be beneficial on certain datasets. Overall, these results show that strong backbone design is a crucial yet previously underappreciated factor in current MLNC benchmarks.

\begin{table*}[t]
  \caption{Main results on five MLNC benchmark datasets. Among prior methods, LANC, ML-GCN, and LARN are selected as the top-ranked non-GCN baselines from the CorGCN benchmark. Lower is better for Ranking Loss and Hamming Loss; higher is better for the remaining metrics. Best results are in \textbf{bold} and second-best results are \underline{underlined}. For Delve, the standard deviation of GCNII is temporarily omitted and will be added in the final version.}
  \label{tab:main_results}
  \centering
  \resizebox{\textwidth}{!}{
  \begin{tabular}{llccccccc}
    \toprule
    Dataset & Metric & LANC & ML-GCN & LARN & CorGCN & GCN & SSGConv & GCNII \\
    \midrule

    \multirow{7}{*}{\rotatebox{90}{Humloc}}
    & Ranking $\downarrow$ & 14.66 $\pm$ 0.36 & 13.28 $\pm$ 0.70 & 13.62 $\pm$ 0.46 & 12.57 $\pm$ 0.31 & \underline{12.31 $\pm$ 0.51} & \textbf{11.98 $\pm$ 0.06} & 12.41 $\pm$ 0.38 \\
    & Hamming $\downarrow$ & 7.82 $\pm$ 0.16 & 7.46 $\pm$ 0.22 & 8.04 $\pm$ 0.29 & \underline{7.37 $\pm$ 0.17} & \textbf{7.35 $\pm$ 0.27} & \textbf{7.35 $\pm$ 0.26} & 7.55 $\pm$ 0.13 \\
    & Ma-AUC $\uparrow$ & 68.73 $\pm$ 1.76 & 72.41 $\pm$ 2.47 & 71.63 $\pm$ 3.30 & 77.31 $\pm$ 1.58 & \textbf{79.35 $\pm$ 2.02} & \underline{78.48 $\pm$ 1.93} & 76.11 $\pm$ 1.81 \\
    & Mi-AUC $\uparrow$ & 86.06 $\pm$ 0.47 & 87.63 $\pm$ 0.52 & 86.63 $\pm$ 0.50 & 88.57 $\pm$ 0.37 & \textbf{89.09 $\pm$ 0.58} & \underline{89.06 $\pm$ 0.31} & 88.63 $\pm$ 0.34 \\
    & Ma-AP $\uparrow$ & 20.94 $\pm$ 1.69 & 23.78 $\pm$ 1.17 & 20.60 $\pm$ 0.34 & 26.91 $\pm$ 1.67 & \textbf{31.05 $\pm$ 1.90} & \underline{29.14 $\pm$ 1.40} & 28.76 $\pm$ 1.50 \\
    & Mi-AP $\uparrow$ & 42.10 $\pm$ 2.61 & 47.76 $\pm$ 1.59 & 41.30 $\pm$ 1.22 & 48.97 $\pm$ 1.30 & \textbf{52.50 $\pm$ 2.03} & \underline{52.06 $\pm$ 1.66} & 49.80 $\pm$ 1.53 \\
    & LRAP $\uparrow$ & 61.02 $\pm$ 1.45 & 64.91 $\pm$ 1.14 & 62.46 $\pm$ 1.46 & 65.40 $\pm$ 0.83 & \underline{67.10 $\pm$ 1.06} & \textbf{67.50 $\pm$ 0.59} & 65.78 $\pm$ 1.37 \\
    \midrule

    \multirow{7}{*}{\rotatebox{90}{PCG}}
    & Ranking $\downarrow$ & 28.26 $\pm$ 0.51 & 27.11 $\pm$ 0.44 & 28.45 $\pm$ 0.53 & \textbf{25.97 $\pm$ 0.57} & \underline{26.84 $\pm$ 0.89} & 27.08 $\pm$ 0.84 & 27.65 $\pm$ 0.39 \\
    & Hamming $\downarrow$ & 12.59 $\pm$ 0.18 & 12.56 $\pm$ 0.31 & 12.65 $\pm$ 0.14 & \textbf{12.41 $\pm$ 0.23} & 12.55 $\pm$ 0.15 & \underline{12.53 $\pm$ 0.15} & 12.55 $\pm$ 0.21 \\
    & Ma-AUC $\uparrow$ & 58.75 $\pm$ 1.05 & 61.48 $\pm$ 0.84 & 57.00 $\pm$ 1.14 & \textbf{64.86 $\pm$ 1.05} & \underline{63.99 $\pm$ 1.00} & 62.97 $\pm$ 1.06 & 62.50 $\pm$ 0.34 \\
    & Mi-AUC $\uparrow$ & 71.31 $\pm$ 0.56 & 72.03 $\pm$ 0.40 & 70.02 $\pm$ 0.47 & \textbf{74.16 $\pm$ 0.61} & \underline{73.20 $\pm$ 0.72} & 72.89 $\pm$ 0.75 & 72.39 $\pm$ 0.41 \\
    & Ma-AP $\uparrow$ & 19.49 $\pm$ 0.85 & 20.95 $\pm$ 0.91 & 16.23 $\pm$ 0.66 & \textbf{24.64 $\pm$ 0.90} & \underline{23.49 $\pm$ 1.24} & 22.09 $\pm$ 1.39 & 22.15 $\pm$ 1.75 \\
    & Mi-AP $\uparrow$ & 27.54 $\pm$ 0.98 & 29.33 $\pm$ 1.21 & 24.87 $\pm$ 0.59 & \textbf{31.91 $\pm$ 1.07} & \underline{30.90 $\pm$ 1.22} & 30.20 $\pm$ 1.63 & 29.96 $\pm$ 2.15 \\
    & LRAP $\uparrow$ & 46.62 $\pm$ 1.27 & 48.21 $\pm$ 0.57 & 45.76 $\pm$ 0.99 & \textbf{49.04 $\pm$ 0.83} & \underline{48.85 $\pm$ 1.20} & 47.94 $\pm$ 1.53 & 47.78 $\pm$ 1.64 \\
    \midrule

    \multirow{7}{*}{\rotatebox{90}{BlogCatalog}}
    & Ranking $\downarrow$ & 25.48 $\pm$ 0.12 & 25.68 $\pm$ 0.25 & 25.67 $\pm$ 0.27 & 25.42 $\pm$ 0.23 & \textbf{23.20 $\pm$ 1.60} & \underline{24.51 $\pm$ 0.79} & 24.77 $\pm$ 0.57 \\
    & Hamming $\downarrow$ & \underline{3.55 $\pm$ 0.03} & 3.58 $\pm$ 0.03 & 3.58 $\pm$ 0.03 & 3.58 $\pm$ 0.03 & \textbf{3.50 $\pm$ 0.07} & \underline{3.55 $\pm$ 0.05} & 3.56 $\pm$ 0.04 \\
    & Ma-AUC $\uparrow$ & 53.29 $\pm$ 0.52 & 50.94 $\pm$ 1.17 & 50.52 $\pm$ 1.07 & 54.48 $\pm$ 0.52 & \textbf{61.52 $\pm$ 5.20} & \underline{57.68 $\pm$ 2.76} & 57.01 $\pm$ 1.71 \\
    & Mi-AUC $\uparrow$ & 74.11 $\pm$ 0.06 & 73.85 $\pm$ 0.17 & 73.73 $\pm$ 0.47 & 74.15 $\pm$ 0.21 & \textbf{76.25 $\pm$ 1.48} & \underline{74.97 $\pm$ 0.64} & 74.73 $\pm$ 0.45 \\
    & Ma-AP $\uparrow$ & 5.07 $\pm$ 0.20 & 4.16 $\pm$ 0.14 & 4.15 $\pm$ 0.48 & 4.61 $\pm$ 0.15 & \textbf{9.93 $\pm$ 3.87} & \underline{6.45 $\pm$ 1.51} & 5.95 $\pm$ 1.27 \\
    & Mi-AP $\uparrow$ & 11.71 $\pm$ 0.55 & 9.42 $\pm$ 0.15 & 9.38 $\pm$ 1.22 & 9.65 $\pm$ 0.21 & \textbf{18.11 $\pm$ 5.89} & \underline{12.92 $\pm$ 2.32} & 11.74 $\pm$ 1.92 \\
    & LRAP $\uparrow$ & 28.94 $\pm$ 1.00 & 27.87 $\pm$ 0.18 & 28.10 $\pm$ 0.92 & 28.32 $\pm$ 0.29 & \textbf{35.08 $\pm$ 4.72} & \underline{31.07 $\pm$ 1.91} & 30.27 $\pm$ 1.59 \\
    \midrule

    \multirow{7}{*}{\rotatebox{90}{PPI}}
    & Ranking $\downarrow$ & 18.44 $\pm$ 0.14 & 20.05 $\pm$ 0.62 & 19.85 $\pm$ 0.18 & 16.17 $\pm$ 0.22 & 13.06 $\pm$ 0.02 & \underline{12.84 $\pm$ 0.06} & \textbf{12.10 $\pm$ 0.17} \\
    & Hamming $\downarrow$ & 21.85 $\pm$ 0.28 & 23.83 $\pm$ 0.14 & 23.41 $\pm$ 0.13 & 20.79 $\pm$ 0.29 & \underline{17.73 $\pm$ 0.09} & 17.91 $\pm$ 0.23 & \textbf{17.04 $\pm$ 0.39} \\
    & Ma-AUC $\uparrow$ & 74.24 $\pm$ 0.42 & 70.26 $\pm$ 1.12 & 70.21 $\pm$ 0.24 & 77.54 $\pm$ 0.42 & 82.88 $\pm$ 0.13 & \underline{83.29 $\pm$ 0.21} & \textbf{84.50 $\pm$ 0.37} \\
    & Mi-AUC $\uparrow$ & 81.15 $\pm$ 0.26 & 78.37 $\pm$ 0.72 & 78.50 $\pm$ 0.13 & 83.35 $\pm$ 0.29 & \underline{87.51 $\pm$ 0.08} & 87.43 $\pm$ 0.19 & \textbf{88.41 $\pm$ 0.27} \\
    & Ma-AP $\uparrow$ & 57.19 $\pm$ 0.15 & 50.69 $\pm$ 1.54 & 51.16 $\pm$ 0.61 & 61.33 $\pm$ 0.67 & 69.54 $\pm$ 0.37 & \underline{70.22 $\pm$ 0.61} & \textbf{72.31 $\pm$ 0.38} \\
    & Mi-AP $\uparrow$ & 69.59 $\pm$ 0.27 & 64.98 $\pm$ 1.32 & 65.57 $\pm$ 0.54 & 72.35 $\pm$ 0.58 & \underline{79.08 $\pm$ 0.29} & 78.99 $\pm$ 0.36 & \textbf{80.53 $\pm$ 0.25} \\
    & LRAP $\uparrow$ & 68.87 $\pm$ 0.35 & 66.68 $\pm$ 1.01 & 67.96 $\pm$ 0.36 & 71.40 $\pm$ 0.34 & 75.34 $\pm$ 0.30 & \underline{75.72 $\pm$ 0.18} & \textbf{76.68 $\pm$ 0.17} \\
    \midrule

    \multirow{7}{*}{\rotatebox{90}{Delve}}
    & Ranking $\downarrow$ & OOM & 3.33 $\pm$ 0.01 & 1.46 $\pm$ 0.02 & 2.41 $\pm$ 0.02 & \underline{1.11 $\pm$ 0.02} & \underline{1.11 $\pm$ 0.04} & \textbf{0.99} \\
    & Hamming $\downarrow$ & OOM & 2.99 $\pm$ 0.01 & 1.67 $\pm$ 0.07 & 2.48 $\pm$ 0.01 & \underline{1.58 $\pm$ 0.01} & \underline{1.58 $\pm$ 0.01} & \textbf{1.53} \\
    & Ma-AUC $\uparrow$ & OOM & 95.71 $\pm$ 0.04 & 97.85 $\pm$ 0.05 & 96.84 $\pm$ 0.10 & 98.51 $\pm$ 0.05 & \underline{98.53 $\pm$ 0.04} & \textbf{98.69} \\
    & Mi-AUC $\uparrow$ & OOM & 97.29 $\pm$ 0.02 & 98.39 $\pm$ 0.07 & 98.05 $\pm$ 0.02 & \underline{99.18 $\pm$ 0.02} & \underline{99.18 $\pm$ 0.02} & \textbf{99.29} \\
    & Ma-AP $\uparrow$ & OOM & 65.20 $\pm$ 0.07 & 80.16 $\pm$ 0.32 & 72.80 $\pm$ 0.26 & 84.23 $\pm$ 0.09 & \underline{84.41 $\pm$ 0.14} & \textbf{85.48} \\
    & Mi-AP $\uparrow$ & OOM & 81.78 $\pm$ 0.09 & 90.44 $\pm$ 0.12 & 86.12 $\pm$ 0.03 & \underline{93.72 $\pm$ 0.05} & \underline{93.75 $\pm$ 0.10} & \textbf{94.31} \\
    & LRAP $\uparrow$ & OOM & 86.12 $\pm$ 0.05 & 92.09 $\pm$ 0.21 & 89.20 $\pm$ 0.01 & \underline{94.58 $\pm$ 0.01} & \underline{94.62 $\pm$ 0.13} & \textbf{95.02} \\
    \bottomrule
  \end{tabular}}
\end{table*}

\subsection{Effect of Backbone Strengthening}

We next examine how much improvement can be obtained by strengthening the standard GCN backbone. Table~\ref{tab:gcn_ablation} compares basic GCN and the strengthened GCN with residual connections and normalization. A consistent pattern can be observed across datasets: adding these simple design choices substantially improves the performance of GCN on most metrics, often narrowing or even reversing the gap with specialized MLNC methods. This result indicates that the apparent weakness of classic GNNs in MLNC may be partly due to under-tuned or under-strengthened backbones rather than an inherent limitation of full-graph message passing itself.

\begin{table}[t]
  \caption{Effect of strengthening the GCN backbone. We compare Basic GCN and Enhanced GCN on five benchmark datasets using three representative metrics. Higher is better for all reported metrics. Mean and standard deviation are reported over multiple runs.}
  \label{tab:gcn_ablation}
  \centering
  \small
  \begin{tabular*}{\columnwidth}{@{\extracolsep{\fill}}llccc}
    \toprule
    Dataset & Model & Ma-AUC $\uparrow$ & Ma-AP $\uparrow$ & LRAP $\uparrow$ \\
    \midrule
    \multirow{2}{*}{Humloc}
      & Basic GCN    & 67.74 $\pm$ 2.35 & 23.08 $\pm$ 0.94 & 63.05 $\pm$ 0.57 \\
      & Enhanced GCN & 79.35 $\pm$ 2.02 & 31.05 $\pm$ 1.90 & 67.10 $\pm$ 1.06 \\
    \midrule
    \multirow{2}{*}{PPI}
      & Basic GCN    & 73.33 $\pm$ 4.80 & 55.08 $\pm$ 6.02 & 69.25 $\pm$ 3.39 \\
      & Enhanced GCN & 82.88 $\pm$ 0.13 & 69.54 $\pm$ 0.37 & 75.34 $\pm$ 0.30 \\
    \midrule
    \multirow{2}{*}{BlogCatalog}
      & Basic GCN    & 62.21 $\pm$ 0.52 & 8.72 $\pm$ 0.31 & 34.79 $\pm$ 0.71 \\
      & Enhanced GCN & 61.52 $\pm$ 5.20 & 9.93 $\pm$ 3.87 & 35.08 $\pm$ 4.72 \\
    \midrule
    \multirow{2}{*}{PCG}
      & Basic GCN    & 65.07 $\pm$ 1.10 & 23.75 $\pm$ 0.09 & 49.19 $\pm$ 0.84 \\
      & Enhanced GCN & 63.99 $\pm$ 1.00 & 23.49 $\pm$ 1.24 & 48.85 $\pm$ 1.20 \\
    \midrule
    \multirow{2}{*}{Delve}
      & Basic GCN    & 92.73 $\pm$ 0.20 & 56.80 $\pm$ 0.36 & 80.65 $\pm$ 0.15 \\
      & Enhanced GCN & 98.51 $\pm$ 0.05 & 84.23 $\pm$ 0.09 & 94.58 $\pm$ 0.01 \\
    \bottomrule
  \end{tabular*}
\end{table}

\begin{table*}[t]
    \centering
    \caption{Component ablation on five benchmark datasets for GCN. We report the effect of removing different strengthening components from Enhanced GCN. Higher is better for all reported metrics. Mean and standard deviation are reported over multiple runs. The Delve result of \textit{w/o Residual} is temporarily unavailable and will be added later.}
    \label{tab:component_ablation}
    \resizebox{\textwidth}{!}{
    \begin{tabular}{llccc}
        \toprule
        \textbf{Dataset} & \textbf{Model Variant} & \textbf{Macro-AUC} $\uparrow$ & \textbf{Macro-AP} $\uparrow$ & \textbf{LRAP} $\uparrow$ \\
        \midrule

        \multirow{5}{*}{Humloc}
            & Basic GCN (w/o Res/Norm/Dropout) & 67.74 $\pm$ 2.35 & 23.08 $\pm$ 0.94 & 63.05 $\pm$ 0.57 \\
            & w/o Dropout (with Res/Norm)      & 73.88 $\pm$ 2.65 & 24.87 $\pm$ 1.23 & 65.11 $\pm$ 2.25 \\
            & w/o Residual (with Norm/Dropout) & 78.10 $\pm$ 1.80 & 26.31 $\pm$ 1.50 & 64.82 $\pm$ 1.03 \\
            & w/o Norm (with Res/Dropout)      & 76.83 $\pm$ 1.75 & 28.51 $\pm$ 1.37 & 65.72 $\pm$ 1.10 \\
            & Enhanced GCN (Full)              & \textbf{79.35 $\pm$ 2.02} & \textbf{31.05 $\pm$ 1.90} & \textbf{67.10 $\pm$ 1.06} \\
        \midrule

        \multirow{5}{*}{PPI}
            & Basic GCN (w/o Res/Norm/Dropout) & 73.33 $\pm$ 4.80 & 55.08 $\pm$ 6.02 & 69.25 $\pm$ 3.39 \\
            & w/o Dropout (with Res/Norm)      & \textbf{91.00 $\pm$ 0.32} & \textbf{82.29 $\pm$ 0.64} & \textbf{85.85 $\pm$ 0.15} \\
            & w/o Residual (with Norm/Dropout) & 80.60 $\pm$ 0.45 & 66.07 $\pm$ 0.42 & 73.89 $\pm$ 0.33 \\
            & w/o Norm (with Res/Dropout)      & 79.97 $\pm$ 0.19 & 64.53 $\pm$ 0.48 & 72.51 $\pm$ 0.34 \\
            & Enhanced GCN (Full)              & 85.44 $\pm$ 0.42 & 73.48 $\pm$ 0.98 & 77.58 $\pm$ 0.69 \\
        \midrule

        \multirow{5}{*}{BlogCatalog}
            & Basic GCN (w/o Res/Norm/Dropout) & \textbf{62.21 $\pm$ 0.52} & 8.72 $\pm$ 0.31 & 34.79 $\pm$ 0.71 \\
            & w/o Dropout (with Res/Norm)      & 56.44 $\pm$ 0.49 & 4.99 $\pm$ 0.23 & 28.77 $\pm$ 0.12 \\
            & w/o Residual (with Norm/Dropout) & 57.02 $\pm$ 7.06 & 7.30 $\pm$ 4.11 & 31.83 $\pm$ 5.27 \\
            & w/o Norm (with Res/Dropout)      & 60.98 $\pm$ 3.40 & 9.07 $\pm$ 1.71 & 34.89 $\pm$ 2.51 \\
            & Enhanced GCN (Full)              & 61.52 $\pm$ 5.20 & \textbf{9.93 $\pm$ 3.87} & \textbf{35.08 $\pm$ 4.72} \\
        \midrule

        \multirow{5}{*}{PCG}
            & Basic GCN (w/o Res/Norm/Dropout) & 65.07 $\pm$ 1.10 & 23.75 $\pm$ 0.09 & 49.19 $\pm$ 0.84 \\
            & w/o Dropout (with Res/Norm)      & 62.07 $\pm$ 1.03 & 21.57 $\pm$ 1.12 & 47.22 $\pm$ 1.17 \\
            & w/o Residual (with Norm/Dropout) & \textbf{66.24 $\pm$ 0.38} & \textbf{25.05 $\pm$ 0.26} & 49.24 $\pm$ 0.89 \\
            & w/o Norm (with Res/Dropout)      & 64.41 $\pm$ 0.58 & 24.72 $\pm$ 0.27 & \textbf{49.52 $\pm$ 0.60} \\
            & Enhanced GCN (Full)              & 64.06 $\pm$ 1.04 & 24.05 $\pm$ 0.69 & 48.97 $\pm$ 1.17 \\
        \midrule

        \multirow{5}{*}{Delve}
            & Basic GCN (w/o Res/Norm/Dropout) & 92.73 $\pm$ 0.20 & 56.80 $\pm$ 0.36 & 80.65 $\pm$ 0.15 \\
            & w/o Dropout (with Res/Norm)      & 98.41 $\pm$ 0.06 & 82.68 $\pm$ 0.17 & 94.26 $\pm$ 0.09 \\
            & w/o Residual (with Norm/Dropout) & -- & -- & -- \\
            & w/o Norm (with Res/Dropout)      & \textbf{98.78 $\pm$ 0.04} & \textbf{85.72 $\pm$ 0.34} & \textbf{94.97 $\pm$ 0.06} \\
            & Enhanced GCN (Full)              & 98.51 $\pm$ 0.05 & 84.23 $\pm$ 0.09 & 94.58 $\pm$ 0.01 \\
        \bottomrule
    \end{tabular}}
\end{table*}

\subsection{Influence of Key Design Factors}

To better understand the source of performance gains, we further analyze the influence of normalization, dropout, residual connections, and network depth.

\paragraph{Normalization and residual connections.}
Normalization and residual connections together play a central role in making classic GNNs competitive. Normalization stabilizes optimization and reduces hidden feature drift across layers, while residual connections preserve useful information from earlier representations and facilitate gradient propagation. Their combination is especially effective when deeper message passing is required.

\paragraph{Dropout.}
Dropout consistently improves generalization by reducing co-adaptation among hidden units. Although its effect is smaller than that of normalization or residual connections in some settings, it remains an important regularization component for obtaining strong and stable performance.



\subsection{Efficiency Analysis}

Table~\ref{tab:efficiency} reports the efficiency comparison on the Humloc dataset. We observe that the strengthened full-graph GNN backbones are consistently more efficient than the more complex architecture in terms of both computation and memory usage. Specifically, CorGCN requires substantially higher training and inference time, as well as larger GPU and CPU memory consumption, while GCN, SSGConv, and GCNII all achieve much lower overhead.

\begin{table}[t]
  \caption{Efficiency comparison on the Humloc dataset. We report the average training time per epoch, inference time, peak GPU memory consumption, and CPU RSS over multiple runs. Lower is better for time and memory metrics.}
  \label{tab:efficiency}
  \centering
  \small
  \begin{tabular*}{\columnwidth}{@{\extracolsep{\fill}}lcccc}
    \toprule
    Model & Train / Epoch (ms) & Infer. (ms) & GPU Peak (MB) & CPU RSS (MB) \\
    \midrule
    CorGCN  & 410.0 & 270.0 & 227.3 & 1372.6 \\
    GCN     & \textbf{109.0} & \textbf{103.4} & 105.2 & 1182.8 \\
    SSGConv & 123.9 & 110.8 & 107.7 & 1186.9 \\
    GCNII   & 116.3 & 105.0 & \textbf{105.0} & \textbf{1178.3} \\
    \bottomrule
  \end{tabular*}
\end{table}

More importantly, this efficiency advantage does not come at the cost of predictive performance. As shown in the main results, these strengthened GNN backbones are not only significantly cheaper to train and deploy, but also achieve the best overall performance on current benchmarks. This demonstrates the practical advantage of our backbone-centric strategy: compared with more complex label-aware architectures such as CorGCN, carefully strengthened full-graph GNNs offer a more favorable trade-off between effectiveness and efficiency.

\section{Discussion and Conclusion}

The results in this work suggest that the performance of multi-label node classification models may depend more strongly on the underlying GNN backbone than previously recognized. Recent studies in MLNC have mainly focused on designing increasingly complex label-aware components, such as label-specific graph views, inter-label correlation modeling, or specialized propagation mechanisms. While these designs are intuitively appealing, our experiments show that, on current benchmark datasets, properly strengthened full-graph GNNs can already achieve highly competitive and often state-of-the-art performance. This indicates that part of the gains previously attributed to task-specific modules may in fact come from the general representation and optimization capacity of the backbone itself.

A key implication of our findings is that strong baseline evaluation is particularly important for MLNC. If the underlying GNN is under-tuned or under-strengthened, then more sophisticated methods may appear to provide larger gains than they actually do. In contrast, once simple yet effective design choices such as normalization, dropout, residual connections, and proper depth tuning are incorporated, classic GNNs become much more competitive. This observation is consistent with recent evidence in standard node classification and suggests that the MLNC setting should also place greater emphasis on fair architectural comparison and careful hyperparameter design, rather than relying solely on increasingly elaborate task-specific modules.

At the same time, our results do not imply that label-aware designs are universally unnecessary. In particular, on PCG, CorGCN still achieves the best performance, which suggests that explicit modeling of label-aware structure and inter-label dependency may remain beneficial for certain datasets. This also points to an interesting direction for future work: rather than asking whether label-aware modules are always useful, it may be more meaningful to study \emph{when} they are useful, and what graph or label-space characteristics make them necessary. Such analysis could help clarify the boundary between improvements brought by stronger backbones and those brought by genuinely task-specific inductive biases.

In summary, this paper provides a systematic reassessment of strong full-graph GNN baselines for multi-label node classification. By revisiting classic backbones such as GCN, SSGConv, and GCNII under simple but important architectural and training enhancements, we show that strong full-graph GNNs already constitute highly competitive baselines on current MLNC benchmarks, surpassing a recent representative specialized method on four out of five datasets. These findings highlight backbone strength as a central yet underappreciated factor in MLNC, and emphasize the importance of fair, rigorous, and informative baseline evaluation for future research in multi-label graph learning.

\bibliographystyle{plainnat}
\bibliography{ref}

\appendix
\subsection{Additional Component Ablation on SSGConv and GCNII}
\label{app:ablation_ssg_gcnii}

To further examine whether the observations made for GCN also generalize to other strong full-graph backbones, we additionally conduct component ablation studies on \textbf{SSGConv} and \textbf{GCNII}. For each backbone, we remove one strengthening component at a time while keeping the remaining designs unchanged. Specifically, we report the variants \textit{w/o Dropout}, \textit{w/o Residual}, and \textit{w/o Norm}, and compare them with the corresponding full model. Following the main text, we focus on three representative metrics, namely \textbf{Macro-AUC}, \textbf{Macro-AP}, and \textbf{LRAP}. The datasets are presented in the order of \textbf{Humloc}, \textbf{PCG}, \textbf{BlogCatalog}, and \textbf{PPI}.

Table~\ref{tab:ssg_component_ablation_appendix} shows the ablation results for SSGConv. Overall, the full SSGConv model performs best on \textbf{Humloc}, \textbf{BlogCatalog}, and \textbf{PPI}, while on \textbf{PCG}, removing residual connections or normalization can even yield stronger results than the full model. This again suggests that the effect of backbone-strengthening components is not uniform across datasets. In particular, residual connections appear consistently beneficial for SSGConv on most datasets, but their contribution becomes less stable on PCG.

Table~\ref{tab:gcnii_component_ablation_appendix} reports the corresponding ablation results for GCNII. Compared with GCN and SSGConv, GCNII exhibits a more dataset-dependent response to these components. On \textbf{Humloc}, removing residual connections or normalization can slightly improve over the full model, while on \textbf{BlogCatalog} and \textbf{PCG}, the \textit{w/o Norm} and \textit{w/o Residual} variants are also competitive or even better. On \textbf{PPI}, the \textit{w/o Dropout} variant achieves the strongest performance among all reported GCNII settings. These results indicate that deeper backbones such as GCNII may require more careful per-dataset configuration, and that ``adding all strengthening components'' is not always optimal.

Overall, these additional experiments reinforce the main conclusion of this paper: the effectiveness of backbone-strengthening strategies is highly \emph{backbone-dependent} and \emph{dataset-dependent}. While normalization, dropout, and residual connections are often beneficial, their gains are not universal, and careful empirical validation remains essential for fair and informative benchmarking in multi-label node classification.

\begin{table*}[t]
    \centering
    \small
    \setlength{\tabcolsep}{8pt}
    \renewcommand{\arraystretch}{1.15}
    \caption{Component ablation on SSGConv. Higher is better for all reported metrics. Mean and standard deviation are reported over multiple runs.}
    \label{tab:ssg_component_ablation_appendix}
    \begin{tabular}{llccc}
        \toprule
        \textbf{Dataset} & \textbf{Model Variant} & \textbf{Macro-AUC} $\uparrow$ & \textbf{Macro-AP} $\uparrow$ & \textbf{LRAP} $\uparrow$ \\
        \midrule
        \multirow{4}{*}{Humloc}
            & w/o Dropout (with Res/Norm)      & 73.13 $\pm$ 2.61 & 24.57 $\pm$ 1.15 & 64.78 $\pm$ 0.78 \\
            & w/o Residual (with Norm/Dropout) & 77.64 $\pm$ 1.92 & 26.24 $\pm$ 1.89 & 64.68 $\pm$ 0.92 \\
            & w/o Norm (with Res/Dropout)      & 75.42 $\pm$ 1.89 & 26.34 $\pm$ 0.23 & 66.51 $\pm$ 1.44 \\
            & Full SSGConv                     & \textbf{78.48 $\pm$ 1.93} & \textbf{29.14 $\pm$ 1.40} & \textbf{67.50 $\pm$ 0.59} \\
        \midrule
        \multirow{4}{*}{PCG}
            & w/o Dropout (with Res/Norm)      & 60.46 $\pm$ 1.22 & 20.10 $\pm$ 0.62 & 47.26 $\pm$ 1.28 \\
            & w/o Residual (with Norm/Dropout) & \textbf{65.58 $\pm$ 0.76} & \textbf{24.26 $\pm$ 1.22} & \textbf{48.81 $\pm$ 0.91} \\
            & w/o Norm (with Res/Dropout)      & 64.14 $\pm$ 1.63 & 23.72 $\pm$ 0.72 & 48.60 $\pm$ 0.89 \\
            & Full SSGConv                     & 62.97 $\pm$ 1.06 & 22.09 $\pm$ 1.39 & 47.94 $\pm$ 1.53 \\
        \midrule
        \multirow{4}{*}{BlogCatalog}
            & w/o Dropout (with Res/Norm)      & \textbf{61.92 $\pm$ 5.26} & \textbf{9.55 $\pm$ 3.54} & \textbf{35.64 $\pm$ 5.03} \\
            & w/o Residual (with Norm/Dropout) & 54.66 $\pm$ 4.68 & 5.56 $\pm$ 1.87 & 29.81 $\pm$ 2.21 \\
            & w/o Norm (with Res/Dropout)      & 55.54 $\pm$ 1.12 & 5.17 $\pm$ 0.52 & 28.90 $\pm$ 0.63 \\
            & Full SSGConv                     & 57.68 $\pm$ 2.76 & 6.45 $\pm$ 1.51 & 31.07 $\pm$ 1.91 \\
        \midrule
        \multirow{4}{*}{PPI}
            & w/o Dropout (with Res/Norm)      & \textbf{91.49 $\pm$ 0.34} & \textbf{82.98 $\pm$ 0.67} & \textbf{86.10 $\pm$ 0.34} \\
            & w/o Residual (with Norm/Dropout) & 77.73 $\pm$ 0.52 & 61.67 $\pm$ 0.58 & 71.93 $\pm$ 0.49 \\
            & w/o Norm (with Res/Dropout)      & 77.35 $\pm$ 0.26 & 60.78 $\pm$ 0.67 & 70.61 $\pm$ 0.44 \\
            & Full SSGConv                     & 83.29 $\pm$ 0.21 & 70.22 $\pm$ 0.61 & 75.72 $\pm$ 0.18 \\
        \midrule
        \multirow{4}{*}{Delve}
            & w/o Dropout (with Res/Norm)      & 98.27 $\pm$ 0.03 & 82.65 $\pm$ 0.26 & 93.94 $\pm$ 0.11 \\
            & w/o Residual (with Norm/Dropout) & 91.48 $\pm$ 0.14 & 54.53 $\pm$ 0.23 & 78.63 $\pm$ 0.07 \\
            & w/o Norm (with Res/Dropout)      & 98.32 $\pm$ 0.05 & 83.68 $\pm$ 0.32 & 94.18 $\pm$ 0.10 \\
            & Full SSGConv                     & \textbf{98.53 $\pm$ 0.04} & \textbf{84.41 $\pm$ 0.14} & \textbf{94.62 $\pm$ 0.13} \\
        \bottomrule
    \end{tabular}
\end{table*}

\begin{table*}[t]
    \centering
    \small
    \setlength{\tabcolsep}{8pt}
    \renewcommand{\arraystretch}{1.15}
    \caption{Component ablation on GCNII. Higher is better for all reported metrics. Mean and standard deviation are reported over multiple runs. For the full GCNII model on Delve, the standard deviation is temporarily unavailable.}
    \label{tab:gcnii_component_ablation_appendix}
    \begin{tabular}{llccc}
        \toprule
        \textbf{Dataset} & \textbf{Model Variant} & \textbf{Macro-AUC} $\uparrow$ & \textbf{Macro-AP} $\uparrow$ & \textbf{LRAP} $\uparrow$ \\
        \midrule
        \multirow{4}{*}{Humloc}
            & w/o Dropout (with Res/Norm)      & 74.00 $\pm$ 2.79 & 24.65 $\pm$ 0.82 & 64.12 $\pm$ 1.42 \\
            & w/o Residual (with Norm/Dropout) & \textbf{78.61 $\pm$ 1.15} & 28.09 $\pm$ 0.13 & \textbf{66.55 $\pm$ 0.91} \\
            & w/o Norm (with Res/Dropout)      & 76.56 $\pm$ 2.16 & \textbf{29.62 $\pm$ 1.47} & 66.41 $\pm$ 1.69 \\
            & Full GCNII                       & 76.11 $\pm$ 1.81 & 28.76 $\pm$ 1.50 & 65.78 $\pm$ 1.37 \\
        \midrule
        \multirow{4}{*}{PCG}
            & w/o Dropout (with Res/Norm)      & 60.82 $\pm$ 0.94 & 19.77 $\pm$ 0.82 & 47.22 $\pm$ 0.87 \\
            & w/o Residual (with Norm/Dropout) & \textbf{65.44 $\pm$ 0.70} & \textbf{24.89 $\pm$ 1.13} & \textbf{49.45 $\pm$ 1.19} \\
            & w/o Norm (with Res/Dropout)      & 62.04 $\pm$ 1.16 & 22.99 $\pm$ 1.03 & 48.76 $\pm$ 0.92 \\
            & Full GCNII                       & 62.50 $\pm$ 0.34 & 22.15 $\pm$ 1.75 & 47.78 $\pm$ 1.64 \\
        \midrule
        \multirow{4}{*}{BlogCatalog}
            & w/o Dropout (with Res/Norm)      & 56.49 $\pm$ 0.43 & 5.12 $\pm$ 0.28 & 28.85 $\pm$ 0.31 \\
            & w/o Residual (with Norm/Dropout) & 52.89 $\pm$ 0.35 & 4.27 $\pm$ 0.14 & 28.16 $\pm$ 0.14 \\
            & w/o Norm (with Res/Dropout)      & \textbf{57.90 $\pm$ 1.02} & \textbf{6.72 $\pm$ 0.61} & \textbf{31.21 $\pm$ 0.33} \\
            & Full GCNII                       & 57.01 $\pm$ 1.71 & 5.95 $\pm$ 1.27 & 30.27 $\pm$ 1.59 \\
        \midrule
        \multirow{4}{*}{PPI}
            & w/o Dropout (with Res/Norm)      & \textbf{92.09 $\pm$ 0.15} & \textbf{84.43 $\pm$ 0.35} & \textbf{86.81 $\pm$ 0.14} \\
            & w/o Residual (with Norm/Dropout) & 83.63 $\pm$ 0.10 & 70.46 $\pm$ 0.25 & 76.01 $\pm$ 0.30 \\
            & w/o Norm (with Res/Dropout)      & 85.64 $\pm$ 0.08 & 74.00 $\pm$ 0.19 & 77.44 $\pm$ 0.25 \\
            & Full GCNII                       & 84.50 $\pm$ 0.37 & 72.31 $\pm$ 0.38 & 76.68 $\pm$ 0.17 \\
        \midrule
        \multirow{4}{*}{Delve}
            & w/o Dropout (with Res/Norm)      & 98.37 $\pm$ 0.06 & 82.92 $\pm$ 0.44 & 94.10 $\pm$ 0.10 \\
            & w/o Residual (with Norm/Dropout) & \textbf{98.96 $\pm$ 0.02} & \textbf{86.09 $\pm$ 0.25} & \textbf{95.34 $\pm$ 0.03} \\
            & w/o Norm (with Res/Dropout)      & 98.57 $\pm$ 0.05 & 84.93 $\pm$ 0.30 & 94.59 $\pm$ 0.08 \\
            & Full GCNII                       & 98.65 $\pm$ 0.07 & 85.35 $\pm$ 0.22 & 94.85 $\pm$ 0.15 \\
        \bottomrule
    \end{tabular}
\end{table*}

\newpage
\section*{NeurIPS Paper Checklist}

The checklist is designed to encourage best practices for responsible machine learning research, addressing issues of reproducibility, transparency, research ethics, and societal impact. Do not remove the checklist: {\bf The papers not including the checklist will be desk rejected.} The checklist should follow the references and follow the (optional) supplemental material.  The checklist does NOT count towards the page
limit. 

Please read the checklist guidelines carefully for information on how to answer these questions. For each question in the checklist:
\begin{itemize}
    \item You should answer \answerYes{}, \answerNo{}, or \answerNA{}.
    \item \answerNA{} means either that the question is Not Applicable for that particular paper or the relevant information is Not Available.
    \item Please provide a short (1--2 sentence) justification right after your answer (even for \answerNA). 
\end{itemize}

{\bf The checklist answers are an integral part of your paper submission.} They are visible to the reviewers, area chairs, senior area chairs, and ethics reviewers. You will also be asked to include it (after eventual revisions) with the final version of your paper, and its final version will be published with the paper.

The reviewers of your paper will be asked to use the checklist as one of the factors in their evaluation. While \answerYes{} is generally preferable to \answerNo{}, it is perfectly acceptable to answer \answerNo{} provided a proper justification is given (e.g., error bars are not reported because it would be too computationally expensive'' or ``we were unable to find the license for the dataset we used''). In general, answering \answerNo{} or \answerNA{} is not grounds for rejection. While the questions are phrased in a binary way, we acknowledge that the true answer is often more nuanced, so please just use your best judgment and write a justification to elaborate. All supporting evidence can appear either in the main paper or the supplemental material, provided in appendix. If you answer \answerYes{} to a question, in the justification please point to the section(s) where related material for the question can be found.

IMPORTANT, please:
\begin{itemize}
    \item {\bf Delete this instruction block, but keep the section heading ``NeurIPS Paper Checklist"},
    \item  {\bf Keep the checklist subsection headings, questions/answers and guidelines below.}
    \item {\bf Do not modify the questions and only use the provided macros for your answers}.
\end{itemize}


\begin{enumerate}

\item {\bf Claims}
    \item[] Question: Do the main claims made in the abstract and introduction accurately reflect the paper's contributions and scope?
    \item[] Answer: \answerTODO{} 
    \item[] Justification: \justificationTODO{}

\item {\bf Limitations}
    \item[] Question: Does the paper discuss the limitations of the work performed by the authors?
    \item[] Answer: \answerTODO{} 
    \item[] Justification: \justificationTODO{}

\item {\bf Theory assumptions and proofs}
    \item[] Question: For each theoretical result, does the paper provide the full set of assumptions and a complete (and correct) proof?
    \item[] Answer: \answerTODO{} 
    \item[] Justification: \justificationTODO{}

    \item {\bf Experimental result reproducibility}
    \item[] Question: Does the paper fully disclose all the information needed to reproduce the main experimental results of the paper to the extent that it affects the main claims and/or conclusions of the paper (regardless of whether the code and data are provided or not)?
    \item[] Answer: \answerTODO{} 
    \item[] Justification: \justificationTODO{}

\item {\bf Open access to data and code}
    \item[] Question: Does the paper provide open access to the data and code, with sufficient instructions to faithfully reproduce the main experimental results, as described in supplemental material?
    \item[] Answer: \answerTODO{} 
    \item[] Justification: \justificationTODO{}

\item {\bf Experimental setting/details}
    \item[] Question: Does the paper specify all the training and test details (e.g., data splits, hyperparameters, how they were chosen, type of optimizer) necessary to understand the results?
    \item[] Answer: \answerTODO{} 
    \item[] Justification: \justificationTODO{}

\item {\bf Experiment statistical significance}
    \item[] Question: Does the paper report error bars suitably and correctly defined or other appropriate information about the statistical significance of the experiments?
    \item[] Answer: \answerTODO{} 
    \item[] Justification: \justificationTODO{}

\item {\bf Experiments compute resources}
    \item[] Question: For each experiment, does the paper provide sufficient information on the computer resources (type of compute workers, memory, time of execution) needed to reproduce the experiments?
    \item[] Answer: \answerTODO{} 
    \item[] Justification: \justificationTODO{}
    
\item {\bf Code of ethics}
    \item[] Question: Does the research conducted in the paper conform, in every respect, with the NeurIPS Code of Ethics \url{https://neurips.cc/public/EthicsGuidelines}?
    \item[] Answer: \answerTODO{} 
    \item[] Justification: \justificationTODO{}

\item {\bf Broader impacts}
    \item[] Question: Does the paper discuss both potential positive societal impacts and negative societal impacts of the work performed?
    \item[] Answer: \answerTODO{} 
    \item[] Justification: \justificationTODO{}
    
\item {\bf Safeguards}
    \item[] Question: Does the paper describe safeguards that have been put in place for responsible release of data or models that have a high risk for misuse (e.g., pre-trained language models, image generators, or scraped datasets)?
    \item[] Answer: \answerTODO{} 
    \item[] Justification: \justificationTODO{}

\item {\bf Licenses for existing assets}
    \item[] Question: Are the creators or original owners of assets (e.g., code, data, models), used in the paper, properly credited and are the license and terms of use explicitly mentioned and properly respected?
    \item[] Answer: \answerTODO{} 
    \item[] Justification: \justificationTODO{}

\item {\bf New assets}
    \item[] Question: Are new assets introduced in the paper well documented and is the documentation provided alongside the assets?
    \item[] Answer: \answerTODO{} 
    \item[] Justification: \justificationTODO{}

\item {\bf Crowdsourcing and research with human subjects}
    \item[] Question: For crowdsourcing experiments and research with human subjects, does the paper include the full text of instructions given to participants and screenshots, if applicable, as well as details about compensation (if any)? 
    \item[] Answer: \answerTODO{} 
    \item[] Justification: \justificationTODO{}

\item {\bf Institutional review board (IRB) approvals or equivalent for research with human subjects}
    \item[] Question: Does the paper describe potential risks incurred by study participants, whether such risks were disclosed to the subjects, and whether Institutional Review Board (IRB) approvals (or an equivalent approval/review based on the requirements of your country or institution) were obtained?
    \item[] Answer: \answerTODO{} 
    \item[] Justification: \justificationTODO{}

\item {\bf Declaration of LLM usage}
    \item[] Question: Does the paper describe the usage of LLMs if it is an important, original, or non-standard component of the core methods in this research? Note that if the LLM is used only for writing, editing, or formatting purposes and does \emph{not} impact the core methodology, scientific rigor, or originality of the research, declaration is not required.
    \item[] Answer: \answerTODO{} 
    \item[] Justification: \justificationTODO{}

\end{enumerate}

\end{document}